\documentclass{article} 
\usepackage{iclr2026_conference,times}

\usepackage{amsmath,amsfonts,bm}

\def\eqref#1{equation~\ref{#1}}

\def\1{\bm{1}}

\DeclareMathAlphabet{\mathsfit}{\encodingdefault}{\sfdefault}{m}{sl}
\SetMathAlphabet{\mathsfit}{bold}{\encodingdefault}{\sfdefault}{bx}{n}

\usepackage{algorithm}
\usepackage{algorithmic}
\usepackage{threeparttable}
\usepackage[utf8]{inputenc} 
\usepackage[T1]{fontenc}    
\usepackage{hyperref}       
\usepackage{url}            
\usepackage{booktabs}       
\usepackage{amsfonts}       
\usepackage{nicefrac}       
\usepackage{microtype}      
\usepackage{graphicx}
\usepackage{float} 

\usepackage{amsmath}
\usepackage{amsthm}
\usepackage{amssymb}
\usepackage{array}
\usepackage{color}
\usepackage{caption}
\usepackage{mathtools}
\usepackage{mathrsfs}
\usepackage{wrapfig}
\usepackage{multirow}
\usepackage{diagbox}
\usepackage{bm}
\usepackage{tablefootnote}
\usepackage{fancybox}
\usepackage{colortbl}
\usepackage{subcaption}
\usepackage{comment}
\usepackage[table]{xcolor}
\usepackage{pifont,xcolor}

\usepackage{fontawesome5}
\usepackage[ruled,vlined,algo2e]{algorithm2e}

\definecolor{SDEblue}{RGB}{28 58 88}

\definecolor{myred}{HTML}{ff7777}

\hypersetup{
 colorlinks=true,
 urlcolor=magenta,
 citecolor=SDEblue,
}
\usepackage{textcomp}

\usepackage{tcolorbox}
\definecolor{lightroyalblue}{HTML}{F6F8FD}
\definecolor{royalblue}{HTML}{4169E1}
\definecolor{lighterblue}{HTML}{f2fafd}
\newtcolorbox{abox}{colback=lightroyalblue,colframe=black}
\definecolor{LightCyan}{rgb}{.9, .95, 1.}

\definecolor{high}{HTML}{ffe5c6}
\definecolor{lower}{HTML}{FF8787}
\definecolor{blank}{HTML}{F1F3F5}
\definecolor{blank1}{HTML}{e5dbff}

\title{Depth–Semantic Alignment and Affinity-Guided Fusion for Structured Radar Point Cloud Generation}

\author{
    Amjad Hussain\quad
    Wenjie Liu\quad
    Yuchen Tan\quad
    Fuyuan Ai\quad
    Zecheng Li\quad
    Xin Qiu\thanks{Corresponding author.}\quad
Chunyi Song\footnotemark[1]\\
    Zhejiang University, China\\
    \texttt{21934205@zju.edu.cn}\quad \texttt{qiuxinzju@zju.edu.cn}\quad \texttt{ cysong@zju.edu.cn}
}

\iclrfinalcopy 
\begin{document}

\maketitle

\begin{abstract}
Point clouds are an important carrier of three-dimensional spatial information, and their quality directly affects the performance of downstream perception tasks such as object detection and tracking. However, millimeter-wave radar point clouds are typically sparse, noisy, and structurally incomplete. To address these limitations, this paper proposes a multimodal point cloud generation method based on vision-radar fusion. The proposed method leverages image semantic information to impose structural constraints and achieve spatial alignment for radar point clouds, while incorporating a sparse completion strategy to enhance point density and recover missing structures. The generated point clouds are further evaluated in object detection and tracking tasks. Experimental results demonstrate that the proposed method effectively improves point cloud quality and enhances the detection accuracy and robustness of perception models in complex environments, providing a practical solution for multisensor point cloud generation and intelligent perception systems.
\end{abstract}
\section{Introduction}
In recent years, 77 GHz millimeter-wave radar has become an important perception sensor in advanced driver assistance systems owing to its reliable ranging capability, strong environmental adaptability, and relatively low cost~\citep{tan20223,han20234d,wang2024multi}. It has been widely deployed in applications such as automatic emergency braking and adaptive cruise control. Compared with visual sensors, millimeter-wave radar is more robust under low-light and adverse weather conditions and can reliably measure the range, azimuth, and velocity of surrounding objects, providing dependable environmental information for intelligent driving systems~\citep{peng20254d}.

Similar to LiDAR point clouds, millimeter-wave radar point clouds can support a variety of perception tasks, including environmental mapping, object detection, and localization. However, because of hardware limitations and conventional signal-processing pipelines, radar point clouds are typically sparse, noisy, and structurally incomplete~\citep{wan2024point,wu2026diffusion}. On the one hand, the limited vertical field of view restricts the acquisition of complete three-dimensional structures. On the other hand, environmental clutter and multipath propagation often introduce spurious points and further degrade point cloud quality. These limitations become more pronounced in downstream tasks such as object detection, object tracking, and simultaneous localization and mapping, restricting the use of millimeter-wave radar in high-precision perception scenarios.

To improve the quality of radar point clouds, several studies have attempted to optimize radar signal-processing procedures~\citep{fan2024enhancing,biswas2025advanced}. Nevertheless, such methods remain constrained by the limited information contained in millimeter-wave radar signals and often exhibit insufficient adaptability in complex driving environments. In contrast, multisensor fusion provides a promising solution by exploiting the complementary characteristics of radar and visual sensors~\citep{xu2026brims,john2026review}. Millimeter-wave radar offers accurate range and velocity measurements, whereas images provide rich semantic and structural information. Despite this complementarity, most existing vision–radar fusion methods focus primarily on downstream perception tasks, with comparatively limited attention paid to directly improving radar point cloud quality through point cloud generation.

To address these issues, we proposes a vision–radar fusion framework for generating dense and structurally complete millimeter-wave radar point clouds, providing a unified data representation for various intelligent perception tasks. The proposed framework consists of three main stages: radar data preprocessing, cross-modal feature alignment and fusion, and point cloud post-processing. During preprocessing, a Hessian-matrix-based peak enhancement method is introduced to strengthen radar responses and suppress noise. In the feature fusion stage, depth and semantic priors are employed to spatially align visual features with radar measurements, followed by multimodal feature fusion in the bird’s-eye-view space. During post-processing, a graph-based geometric optimization method is applied to impose structural constraints and refine the generated point clouds.

The generated point clouds are further evaluated on object detection and object tracking tasks. Experimental results demonstrate that the proposed method effectively improves the structural completeness and semantic consistency of radar point clouds, while enhancing the robustness and accuracy of perception models in complex environments. This study provides a practical reference for vision–radar point cloud generation and its application in intelligent perception systems.

\section{Related Work}
\subsection{Vision-Based Pseudo-Point Cloud Generation}
Vision-based pseudo-point cloud generation methods typically recover scene geometry through depth estimation and then project image pixels into 3D space using camera parameters~\citep{zhang2025survey,huang2025systematic}. Existing approaches can be broadly categorized into monocular and stereo methods~\citep{zhao2020monocular,ming2021deep}. Monocular methods estimate pixel-wise depth from a single image and have gradually evolved from handcrafted geometric and visual priors to deep learning architectures based on encoder–decoder networks, multi-scale feature fusion, attention mechanisms, and auxiliary semantic supervision~\citep{zhao2020monocular,mohadikar2025sn360,tosi2025survey}. Stereo methods recover depth by estimating disparity between paired images, using either traditional matching pipelines or deep neural networks with learned matching costs, end-to-end disparity regression, and cost-volume construction~\citep{wang2025scalable,liu2025transformer,hu2024tfdepth}. Although these methods have improved pseudo-point cloud generation, they still suffer from scale ambiguity, incomplete geometry, sensitivity to weak-texture and occluded regions, limited cross-scene generalization. 

\subsection{Radar Signal-Based Point Cloud Generation}
Radar signal-based point cloud generation methods recover target range, velocity, and angular information from raw millimeter-wave radar measurements and convert them into 3D points~\citep{qian20203d,gall2020spectrum}. Existing approaches can be broadly divided into array signal processing-based and deep learning-based methods~\citep{zhou2025indoor,hasan2024mm,geng2024dream}. Traditional methods usually employ FMCW signal processing, range–Doppler FFT, CFAR detection, and direction-of-arrival estimation to construct radar point clouds~\citep{cui2023milipoint,katkovnik2002high}. These methods are physically interpretable and stable, but the limited number of antennas often leads to sparse point clouds with incomplete structures and sensitivity to noise and multipath interference. Deep learning methods improve radar point cloud generation by learning features for angle estimation, target detection, denoising, completion, and structural refinement~\citep{cha2021multi,hoang2026deep}. Some recent approaches further reconstruct higher-resolution 3D structures directly from raw radar signals or multi-frame observations. Despite improved robustness and representation capability, these methods generally require high-quality training data and introduce additional computational costs.

\subsection{Vision–Radar Fusion-Based Point Cloud Generation}
Vision–radar fusion-based point cloud generation combines the rich semantic and structural information of images with the accurate range and velocity measurements of millimeter-wave radar~\citep{zhu2025depth,deng2025advances}. Compared with single-sensor approaches, multimodal fusion can alleviate the limitations of insufficient spatial resolution, incomplete depth information, and limited environmental robustness~\citep{nobis2019deep,wang2025c4rfnet}. Existing studies generally follow two directions. Some methods directly generate enhanced radar point clouds and evaluate their geometric quality, although research along this line remains relatively limited. Most approaches instead treat point cloud generation as an intermediate representation for downstream tasks, integrating visual and radar features to improve object detection, object tracking, and other perception tasks~\citep{nabati2021centerfusion,haensel2025robust,bai2021robust}. Despite promising results, achieving accurate cross-modal alignment and generating geometrically consistent point clouds remain important challenges.
\section{Method}
\begin{figure}[t]
    \centering
    \includegraphics[width=0.98\linewidth]{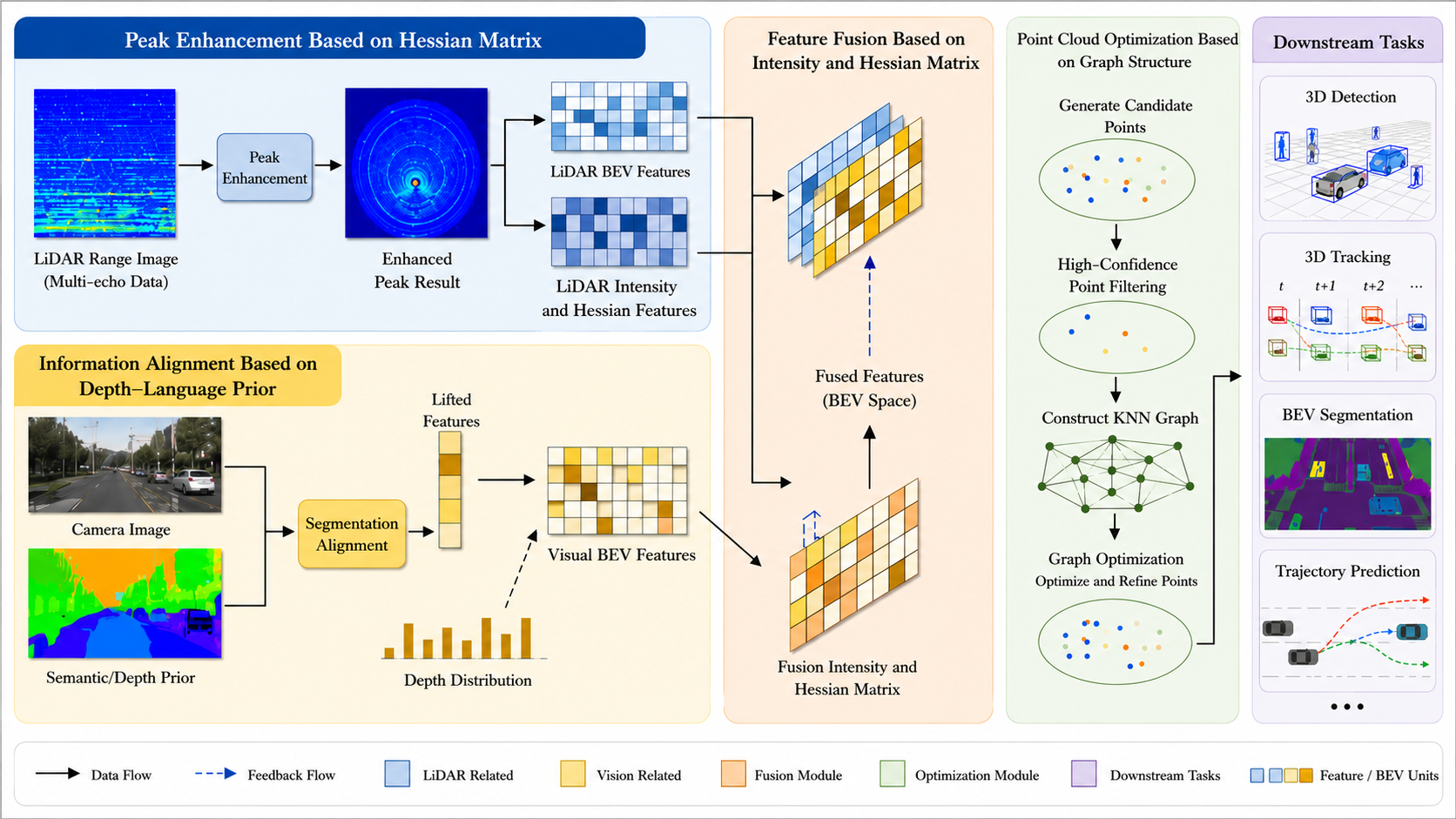}
    \caption{Overview of the proposed method framework. Radar BEV preprocessing with bilinear coordinate transformation, RANSAC-based ground removal, and Hessian-based peak enhancement for improved target responses and boundary clarity.}
    \label{fig:main}
\end{figure}

Our proposed framework is illustrated in Figure~\ref{fig:main}. It generates high-quality millimeter-wave radar point clouds through vision–radar fusion. Radar features are first enhanced in the BEV space using a Hessian-based peak enhancement module, while image features are projected into BEV using depth and semantic priors. Radar-guided feature completion is then applied to improve cross-modal fusion, and the fused representation is decoded into a dense radar point cloud. Finally, graph-based geometric optimization further improves point accuracy and structural completeness.

\subsection{Millimeter-Wave Radar Data Preprocessing}
Range–azimuth–Doppler (RAD) data play an important role in point cloud generation because they preserve raw radar responses before constant false alarm rate (CFAR) processing, thereby avoiding premature removal of target signals. In the BEV space, RAD intensity indicates the spatial likelihood of target presence, with object regions generally exhibiting stronger energy responses than the background.

In the proposed method, bilinear interpolation is first used to transform RAD data from polar coordinates into Cartesian coordinates, producing continuous and uniformly distributed radar BEV features. Meanwhile, LiDAR point clouds are transformed into the radar coordinate system and used as supervision to provide accurate geometric constraints. RANSAC is applied to remove LiDAR ground points and reduce inconsistencies caused by ground reflections. Since radar BEV features often exhibit blurred boundaries around vehicles and pedestrians, a Hessian-based peak enhancement method is further introduced to strengthen target responses and suppress background noise. As shown in Algorithm~\ref{alg:hessian_peak_enhancement}, this preprocessing improves local structural clarity, boundary definition, and feature consistency, providing more reliable inputs for subsequent radar point cloud generation.

\begin{algorithm2e}[t]
\caption{Hessian-Based Peak Enhancement Algorithm}
\label{alg:hessian_peak_enhancement}

\KwIn{Radar BEV grayscale image $I_r(x,y)$}
\KwOut{Enhanced radar BEV image $I_e(x,y)$}

\BlankLine

$I_g(x,y)=I_r(x,y)*G(x,y)$\;

\ForEach{pixel $(x,y)$ in $I_g(x,y)$}{
    \eIf{$I_g(x,y)$ is a peak}{
        $I_e(x,y)=\left|I_g(x,y)*H(x,y)\right|$\;

        $I_{s,p}(x,y)=I_e(x,y)*G(x,y)$\;

        $E_p=\displaystyle\sum_{i,j}I_{s,p}(i,j)^2$\;

        $I_e(x,y)=I_{s,p}\cdot E_p$\;
    }{
        $I_e(x,y)=I_g(x,y)$\;
    }
}

\Return{$I_e(x,y)$}\;

\end{algorithm2e}

\subsubsection{Depth--Semantic Prior-Based Information Alignment}

Similar to the Lift-Splat-Shoot (LSS) framework~\citep{philion2020lift}, let
$X \in \mathbb{R}^{3 \times H \times W}$ denote an RGB image with camera intrinsic matrix $\mathbf{K}$ and extrinsic matrix $\mathbf{E}$. For each image pixel $p=(h,w)$, we associate $D$ discrete depth hypotheses along its viewing ray, forming a set of 3D sampling points:
\begin{equation}
\left\{(h,w,d)\in\mathbb{R}^{3}\mid d\in\mathcal{D}\right\},
\quad
\mathcal{D}=\{\Delta,2\Delta,\ldots,|\mathcal{D}|\Delta\}.
\end{equation}

These sampling points are projected into the 3D space according to the camera geometry, producing a voxel representation of size $D\times H\times W$. Since this lifting process is fully determined by the camera calibration parameters, it introduces no additional learnable parameters.

Conventional LSS methods estimate a global depth distribution, which may overlook category-specific geometry and lead to inaccurate predictions near boundaries or under occlusion and viewpoint changes. We therefore decompose depth estimation into semantic category-specific subproblems to improve visual–radar geometric alignment.

\begin{figure}[t]
    \centering
    \includegraphics[width=0.98\linewidth]{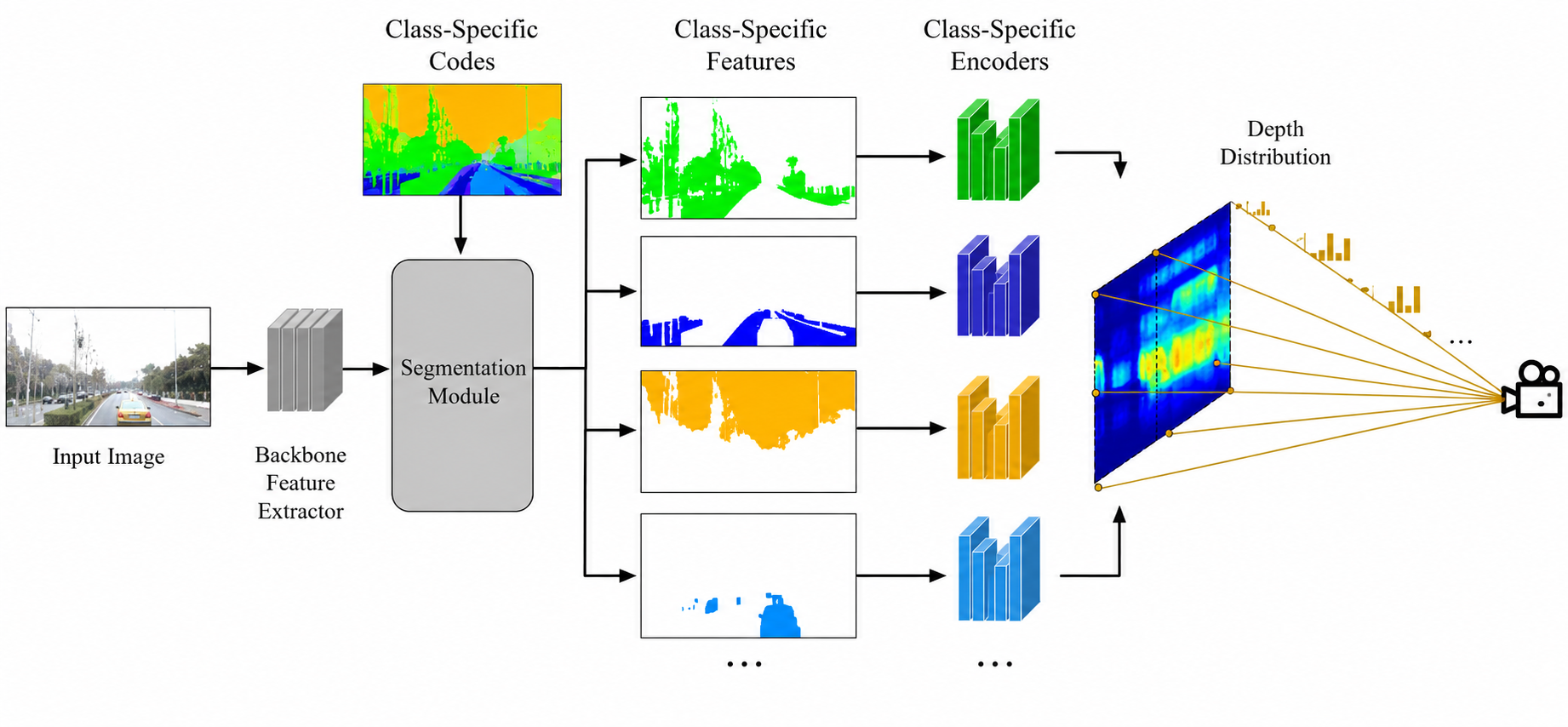}
    \caption{Semantic-based depth estimation framework.}
    \label{fig:Depth}
\end{figure}
As illustrated in Figure~\ref{fig:Depth}, a convolutional network first extracts the guidance feature $\mathbf{G}$ from the input image:
$
\mathbf{G}=\operatorname{CNN}(X),
$
where $\mathbf{G}$ contains both spatial and semantic information. A semantic mask $\mathbf{m}$ is then used to divide $\mathbf{G}$ into nine semantic categories, including traffic facilities, buildings, vehicles, trees, roads, pedestrians, sky, rivers, and tunnels. For the $i$-th category, the corresponding feature is obtained by
\begin{equation}
\mathbf{G}_{i}
=
\operatorname{Decompose}
\left(
\mathbf{G},
\mathbf{m}_{i}
\right),
\quad
i=1,2,\ldots,9.
\end{equation}

Each category-specific feature is processed by an independent LSS encoder to estimate its contextual representation $\mathbf{c}_{i}$ and discrete depth distribution $\boldsymbol{\alpha}_{i}$:
$
\left(
\mathbf{c}_{i},
\boldsymbol{\alpha}_{i}
\right)
=
\operatorname{Encoder}_{i}
\left(
\mathbf{G}_{i}
\right).
$
The category-level outputs are subsequently aggregated to obtain the complete scene context $\mathbf{c}\in\mathbb{R}^{C}$ and depth distribution $\boldsymbol{\alpha}\in\Delta^{D-1}$. For pixel $p$ and depth hypothesis $d$, the lifted feature is defined as
$
\mathbf{c}_{d}
=
\alpha_{d}\mathbf{c},
$
where $\alpha_{d}$ denotes the probability associated with depth $d$. The resulting depth-aware features are then projected into the BEV space using the camera calibration parameters. By incorporating category-specific depth priors, the proposed method improves local depth estimation and provides more reliable geometric representations for subsequent vision--radar alignment and multimodal feature fusion.

\subsubsection{Affinity Matrix-Based Feature Fusion}

The visual BEV features are typically sparse, whereas the radar BEV features exhibit a relatively dense spatial distribution. This imbalance may lead to inaccurate correspondence during cross-modal fusion. To address this issue, we propose an affinity matrix-based feature fusion method that progressively completes the visual BEV representation using the local structural information encoded by radar features.

Given a visual BEV feature map $\mathbf{F}\in\mathbb{R}^{m\times n\times c}$ and the corresponding radar affinity matrix $\mathbf{A}\in\mathbb{R}^{m\times n\times c}$, the visual features are refined through $N$ iterative propagation steps. At iteration $t$, the feature update at position $(i,j)$ is formulated as:
\begin{equation}
\begin{aligned}
\mathbf{F}_{i,j,t+1}
={}
\kappa_{i,j}(0,0)\odot\mathbf{F}_{i,j,0}
+
\sum_{\substack{
a,b=-{(k-1)}/{2}
}}^{{(k-1)}/{2}}
\kappa_{i,j}(a,b)\odot
\mathbf{F}_{i-a,j-b,t},
\end{aligned}
\label{eq:affinity_propagation}
\end{equation}
where $k$ denotes the convolution kernel size, $\odot$ represents element-wise multiplication, and $\kappa_{i,j}(a,b)$ denotes the propagation weight from the neighboring position $(i-a,j-b)$ to $(i,j)$. The propagation weights are derived from the local radar affinity coefficients and normalized as:
\begin{equation}
\kappa_{i,j}(a,b)
=
\frac{
\hat{\kappa}_{i,j}(a,b)
}{
\displaystyle
\sum_{(a,b)\neq(0,0)}
\left|
\hat{\kappa}_{i,j}(a,b)
\right|
},
\label{eq:affinity_normalization}
\end{equation}
while the weight of the central position is defined as:
\begin{equation}
\kappa_{i,j}(0,0)
=
1-
\sum_{(a,b)\neq(0,0)}
\kappa_{i,j}(a,b).
\label{eq:center_weight}
\end{equation}

The coefficient tensor
$\hat{\boldsymbol{\kappa}}_{i,j}\in\mathbb{R}^{k\times k\times c}$
is predicted from the local spatial information of the dense radar BEV features, enabling the propagation process to preserve object structures. An odd kernel size is adopted to maintain spatial symmetry around each pixel. Following the stability constraint~\citep{liu2017learning}, the propagation weights are normalized to $(-1,1)$ when
\begin{equation}
\sum_{(a,b)\neq(0,0)}
\left|
\kappa_{i,j}(a,b)
\right|
\leq 1.
\end{equation}

After $N$ iterations, the completed visual BEV features are concatenated with the radar BEV features. The resulting multimodal representation contains richer geometric and semantic information and is subsequently used to generate the enhanced radar point cloud $\mathbf{P}_{R}$.

\subsection{Point Cloud Post-Processing}

Although vision--radar fusion captures most object structures, radar BEV noise, feature ambiguity, and visual occlusion can still cause missing points and localization errors. To address this issue without increasing computational cost, we use sparse but reliable CFAR-derived radar points as geometric anchors and propose a graph-based optimization method to refine the generated point cloud $\mathbf{P}_{R}$ using the reference point cloud $\mathbf{P}_{L}$.
The reference point cloud $\mathbf{P}_{L}$ provides accurate geometric measurements, while $\mathbf{P}_{R}$ preserves richer structural details. We construct a directed $K$-nearest-neighbor graph over $\mathbf{P}_{R}$ to model local geometry, where each point is connected to its $K$ nearest neighbors. An accelerated KD-tree is used to improve nearest-neighbor search efficiency~\citep{shevtsov2007highly}.

Let
$ 
\mathbf{Z}\in\mathbb{R}^{(n+m)\times 3}
$ 
denote the generated radar point set, where the first $n$ points have corresponding radar reference points and the remaining $m$ points do not. The associated radar reference points are denoted by
$\mathbf{G}\in\mathbb{R}^{n\times 3}$. For each point $i$, let $\mathcal{N}_{i}$ denote its neighborhood, and let
$\mathbf{W}\in\mathbb{R}^{(n+m)\times(n+m)}$ be the graph weight matrix, where $W_{i,j}$ represents the edge weight between points $i$ and $j$. Inspired by manifold learning~\citep{weinberger2005nonlinear}, the weights are determined by reconstructing each point from its local neighbors:
\begin{equation}
\mathbf{W}
=
\arg\min_{\mathbf{W}}
\left\|
\mathbf{Z}-\mathbf{W}\mathbf{Z}
\right\|_{F}^{2},
\quad
\mathrm{s.t.}\quad
\mathbf{W}\mathbf{1}=\mathbf{1},
\quad
W_{i,j}=0\ \mathrm{if}\ j\notin\mathcal{N}_{i},
\label{eq:graph_weight}
\end{equation}
where $\mathbf{1}\in\mathbb{R}^{n+m}$ is an all-one vector. When the points are in general position and $K>3$, multiple solutions may satisfy the reconstruction constraint. We therefore select the minimum-$\ell_{2}$-norm solution to reduce sensitivity to noise.
The optimized point set is denoted by:
\begin{equation}
\mathbf{Z}'
=
\begin{bmatrix}
\mathbf{Z}'_{P_L}\\
\mathbf{Z}'_{P_R}
\end{bmatrix},
\end{equation}
where $\mathbf{Z}'_{P_L}\in\mathbb{R}^{n\times 3}$ corresponds to the aligned reference points and
$\mathbf{Z}'_{P_R}\in\mathbb{R}^{m\times 3}$ represents the remaining generated points. The first $n$ points are constrained by the radar reference points, while the other points are optimized according to the local geometric relationships encoded by $\mathbf{W}$. The final point cloud is obtained by solving
\begin{equation}
\mathbf{Z}'
=
\arg\min_{\mathbf{Z}'}
\left\|
\mathbf{Z}'-\mathbf{W}\mathbf{Z}'
\right\|_{F}^{2},
\quad
\mathrm{s.t.}\quad
\mathbf{Z}'_{1:n}=\mathbf{G}.
\label{eq:point_cloud_optimization}
\end{equation}

This optimization preserves the local geometry of the generated point cloud while aligning it with reliable radar measurements, thereby improving spatial accuracy and structural completeness.

\subsection{Loss Function Design}

All components of the proposed model are differentiable, enabling end-to-end training. The model predicts a position mask
$\mathbf{m}_{\mathrm{pos}}\in\mathbb{R}^{m\times n\times 2}$
and a height range
$\mathbf{h}\in\mathbb{R}^{m\times n\times 2}$.
The position mask determines whether points should be generated at each BEV location, while the height range specifies the minimum and maximum heights for reconstructing the 3D point distribution.
The position mask is optimized using a cross-entropy loss, whereas the height prediction is supervised by the mean absolute error. The overall point generation loss is defined as:
\begin{equation}
\mathcal{L}_{\mathrm{pts}}
=
\operatorname{CrossEntropy}
\left(
\mathbf{m}_{\mathrm{pos}},
\mathbf{m}_{\mathrm{pos}}^{\mathrm{gt}}
\right)
+
\operatorname{MAE}
\left(
\mathbf{h},
\mathbf{h}^{\mathrm{gt}}
\right),
\end{equation}
where $\mathbf{m}_{\mathrm{pos}}^{\mathrm{gt}}$ and
$\mathbf{h}^{\mathrm{gt}}$ denote the position and height supervision labels generated by projecting the LiDAR point cloud into the BEV space.
\section{Results}
\subsection{Evaluation Settings}
We evaluate the quality of the generated radar point clouds on two downstream tasks: 3D object detection and 3D object tracking. The proposed method is compared with OS-CFAR~\citep{gandhi1988analysis}, Sparse2Dense~\citep{ma2018sparse}, and SGDNet~\citep{li2024semantic} under the same evaluation settings. For object detection, SECOND~\citep{yan2018second} and PointPillars~\citep{lang2019pointpillars} are adopted as representative detectors, and performance is measured using AP$_{30}$ and AP$_{50}$ at IoU thresholds of 0.3 and 0.5, respectively. For object tracking, a SECOND-based tracking framework is evaluated using MOTA and AUC. We also provide qualitative comparisons to assess point cloud density, object boundary completeness, geometric consistency, and background noise.

\subsection{Effect of High-Quality Point Clouds on 3D Object Detection}
To evaluate the effectiveness of the generated point clouds, we compare our method with OS-CFAR~\citep{gandhi1988analysis}, Sparse2Dense~\citep{ma2018sparse}, and SGDNet~\citep{li2024semantic} using two representative 3D detectors, SECOND~\citep{yan2018second} and PointPillars~\citep{lang2019pointpillars}. AP$_{30}$ and AP$_{50}$ denote the average precision at IoU thresholds of 0.3 and 0.5, respectively.

\begin{table*}[h]
\centering
\caption{Comparison of 3D object detection performance.}
\resizebox{0.65\textwidth}{!}{
\begin{tabular}{l|cc|cc}
\toprule
\multirow{2}{*}{\textbf{Methods}} 
& \multicolumn{2}{c|}{\textbf{SECOND}} 
& \multicolumn{2}{c}{\textbf{PointPillars}} \\
& \textbf{AP$_{30}\uparrow$} 
& \textbf{AP$_{50}\uparrow$} 
& \textbf{AP$_{30}\uparrow$} 
& \textbf{AP$_{50}\uparrow$} \\
\midrule
OS-CFAR~\citep{gandhi1988analysis}       & 0.2  & 0.0 & 0.2  & 0.0 \\
Sparse2Dense~\citep{ma2018sparse} & 24.1 & 4.8 & 30.9 & 9.6 \\
SGDNet~\citep{li2024semantic}       & 26.3 & 4.9 & 32.0 & 11.7 \\
\midrule
\textbf{Our Method} 
& \textbf{31.9} 
& \textbf{7.7} 
& \textbf{38.0} 
& \textbf{15.0} \\
\bottomrule
\end{tabular}}
\label{tab:detection_results}
\end{table*}

\begin{figure}[t]
    \centering
    \includegraphics[width=0.98\linewidth]{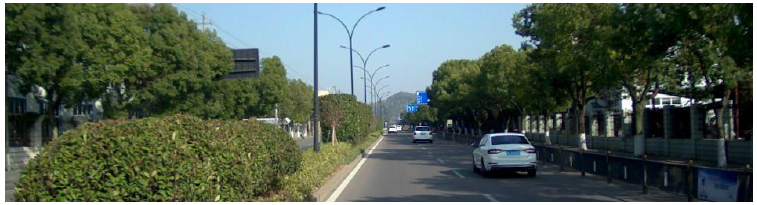}
    \caption{Input RGB image.}
    \label{fig:RGB}
\end{figure}

\begin{figure}[t]
    \centering
    \includegraphics[width=0.98\linewidth]{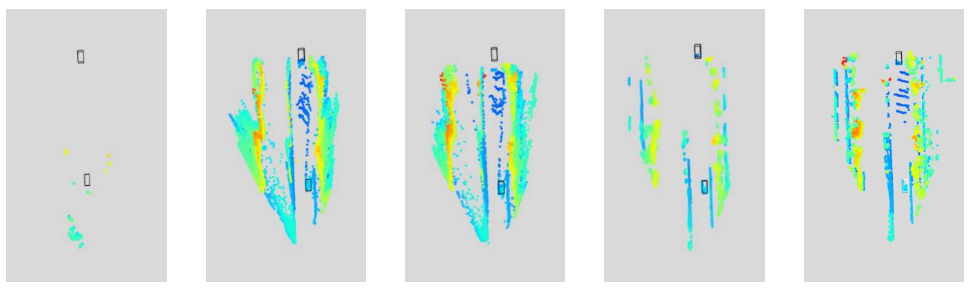}
    \caption{Visual comparison of different point clouds.}
    \label{fig:point}
\end{figure}

As shown in Table~\ref{tab:detection_results}, our method consistently outperforms the competing approaches on both detectors. Compared with SGDNet, it improves AP$_{30}$ and AP$_{50}$ by 5.6 and 2.8 points on SECOND, and by 6.0 and 3.3 points on PointPillars. These results demonstrate that the generated point clouds provide denser structures and more accurate object localization based on Figure~\ref{fig:RGB}.
Figure~\ref{fig:point} further compares the generated point clouds. OS-CFAR produces extremely sparse observations, while Sparse2Dense and SGDNet still contain incomplete boundaries and background noise. In contrast, our method generates denser target-related points and more complete object structures, providing more reliable inputs for 3D detection.

\subsection{Effect of High-Quality Point Clouds on 3D Object Tracking}

High-quality radar point clouds are important for accurate and stable object tracking, particularly under adverse weather, nighttime, long-range, and occlusion conditions. Raw point clouds obtained using CFAR are typically sparse and noisy, which may lead to unstable detections, incorrect cross-frame associations, and fragmented trajectories. To evaluate the effectiveness of the generated point clouds, we use the proposed method as a preprocessing module and feed the enhanced point clouds into a SECOND-based tracking framework.

\begin{table}[h]
\caption{Comparison of 3D object tracking performance.}
\centering
\resizebox{0.7\columnwidth}{!}{
\begin{tabular}{l|cc}
\toprule
\textbf{Methods} 
& \textbf{SECOND-MOTA$\uparrow$} 
& \textbf{SECOND-AUC$\uparrow$} \\
\midrule
OS-CFAR       & 0.108 & 0.112 \\
Sparse2Dense  & 0.412 & 0.503 \\
\midrule
\textbf{Our Method} 
& \textbf{0.494} 
& \textbf{0.572} \\
\bottomrule
\end{tabular}}
\label{tab:tracking_results}
\end{table}

As shown in Table~\ref{tab:tracking_results}, our method achieves the best tracking performance, improving MOTA from 0.108 to 0.494 and AUC from 0.112 to 0.572 compared with OS-CFAR. It also outperforms Sparse2Dense by 0.082 in MOTA and 0.069 in AUC. The denser and more structurally complete point clouds improve object localization and cross-frame association, thereby reducing trajectory fragmentation and identity switches, particularly for distant, small, and partially occluded objects.
\section{Conclusion and Limitations}

This paper presents a vision--radar fusion framework for generating dense and geometrically reliable millimeter-wave radar point clouds. The proposed method enhances radar responses using a Hessian-based preprocessing strategy, improves visual--radar alignment through depth--semantic priors, and completes sparse visual BEV features using radar-guided affinity propagation. A graph-based post-processing module further refines the generated point clouds using reliable CFAR-derived radar points as geometric anchors. Experiments on 3D object detection and tracking demonstrate that the generated point clouds provide more complete object structures, more accurate localization, and improved downstream perception performance compared with conventional radar processing and existing pseudo-point cloud generation methods.

Despite these improvements, several limitations remain. First, the performance of the framework depends on accurate camera--radar calibration, and calibration errors may degrade cross-modal alignment. Second, visual occlusion, severe illumination changes, and inaccurate semantic predictions can affect depth estimation and point cloud generation. Third, the current graph-based refinement and iterative feature propagation introduce additional computational overhead. Future work will investigate calibration-robust fusion, temporal information aggregation, and more efficient point cloud generation to improve generalization and real-time deployment in complex driving environments.
\newpage

\bibliography{iclr2026_conference}
\bibliographystyle{iclr2026_conference}

\end{document}